\documentclass{article}

\usepackage[preprint]{neurips_2024}

\usepackage[T1]{fontenc}
\usepackage[utf8]{inputenc}
\usepackage{amsmath}
\usepackage{amssymb}
\usepackage{graphicx}
\usepackage{multirow,array,makecell}
\usepackage{booktabs}
\usepackage[labelfont=bf,font=small]{caption}
\usepackage{microtype}
\usepackage{url}
\usepackage{hyperref}

\title{uTHCD-C: A Corruption-Robustness Benchmark for Tamil Handwritten Character Recognition}

\author{%
  Preethi S\thanks{Corresponding author: \texttt{preethi.s.ece@kongu.edu}} \\
  Dept.\ of Electronics and Communication Engineering \\
  Kongu Engineering College \\
  Erode, Tamil Nadu, India \\
  \And
  Meeradevi T \\
  Dept.\ of Electronics and Communication Engineering \\
  Kongu Engineering College \\
  Erode, Tamil Nadu, India \\
  \And
  Arun Prakash A \\
  AI4Bharat \\
  Indian Institute of Technology Madras \\
  Chennai, Tamil Nadu, India \\
}

\begin{document}
\maketitle

\begin{abstract}
Evaluation of Tamil handwritten character recognition (HCR) has largely focused on clean test images, and a standardized multi-corruption robustness benchmark for isolated Tamil HCR has not been established. We introduce uTHCD-C, a corruption-robustness benchmark based on the unconstrained Tamil Handwritten Character Database (uTHCD). The benchmark defines ten corruption types, including a handwriting-specific stroke-thinning corruption, at five severity levels. We evaluate nine convolutional and transformer-based architectures using five-fold stratified cross-validation and writer-independent testing. Robustness is quantified using mean Corruption Error (mCE), normalized against a VGG16-BN reference. We additionally compare parameter count, computational complexity, and inference throughput to characterize the computational requirements of the evaluated architectures. The results establish comparative robustness baselines for Tamil HCR, revealing substantial differences across architectures: ConvNeXt and Swin achieve the lowest mCE values (0.75 and 0.83, respectively), whereas ShuffleNet and MnasNet exhibit higher values (1.55 and 1.85, respectively). The accompanying efficiency measurements characterize the trade-offs between corruption robustness and computational requirements. By providing standardized corruption configurations, evaluation protocols, and comparative results, uTHCD-C establishes a reproducible basis for future research on robust and computationally efficient Tamil HCR.
\end{abstract}

\begin{center}
\begin{minipage}{0.9\textwidth}
\end{minipage}
\end{center}

\begin{center}
\small Code and data: \url{https://github.com/pre-prakash/uTHCD-C-Bench}
\end{center}

\bigskip

\section{Introduction}\label{sec1}
Handwritten character recognition (HCR) is an important component of document digitization and language technology. However, recognition accuracy on clean test images does not fully characterize a model's performance when handwriting is affected by image degradation. Scanned and camera-captured documents may contain noise, blur, reduced contrast, spatial distortions, and variations in stroke visibility. Such degradations can alter the structural details that distinguish visually similar characters and reduce recognition performance even when a model performs well under clean conditions.

The unconstrained Tamil Handwritten Character Database (uTHCD) provides a substantial resource for isolated Tamil handwritten-character recognition, comprising 90,950 samples across 156 character classes written by 753 individuals~\cite{Shaffi2021uthcd}. Subsequent Tamil HCR studies have established recognition baselines using conventional evaluation protocols~\cite{kavitha2022new,kavitha2022}. These studies provide important foundations for character recognition, but clean-image evaluation does not systematically establish how model performance changes across controlled corruption types and severity levels. Benchmarks such as ImageNet-C and MNIST-C provide established protocols for evaluating robustness to common corruptions~\cite{imagenet-c,mnist-c}. To the best of our knowledge, no comparable multi-corruption benchmark has been established for isolated Tamil handwritten character recognition.

To address this gap, we introduce \textbf{uTHCD-C}, a corruption-robustness benchmark for isolated Tamil HCR based on uTHCD. The benchmark defines ten corruption types evaluated at five severity levels, including a handwriting-specific stroke-thinning corruption. Nine corruption types are adapted from established corruption-benchmarking conventions, while stroke thinning addresses changes in glyph structure that may not be adequately represented by general image degradations. We evaluate nine convolutional and transformer-based architectures using five-fold stratified cross-validation and writer-independent testing. Robustness is summarized using mean Corruption Error (mCE), normalized against a VGG16-BN reference, alongside clean-data accuracy and corruption-specific results.

We additionally report parameter count, computational complexity, and measured inference throughput alongside the robustness results to support model selection under differing accuracy, robustness, and resource constraints. The evaluation reveals substantial differences in corruption robustness among architectures with similarly high clean-data accuracy: ConvNeXt and Swin achieve the lowest aggregate mCE values, whereas ShuffleNet and MnasNet exhibit substantially higher corruption error. The accompanying efficiency measurements further show that parameter count, computational complexity, and throughput are not interchangeable, exposing distinct robustness--efficiency trade-offs.

The contributions of this study are threefold:
\begin{enumerate}
    \item \textbf{Benchmark construction:} We introduce uTHCD-C, a reproducible multi-corruption benchmark for Tamil handwritten-character recognition, comprising ten corruption types and five severity levels, including a handwriting-specific stroke-thinning corruption.
    \item \textbf{Comparative robustness evaluation:} We establish corruption-robustness baselines for nine recognition architectures using five-fold stratified cross-validation, aggregate mCE, corruption-specific results, and writer-independent testing.
    \item \textbf{Multi-criteria characterization:} We complement the robustness analysis with comparisons of clean recognition accuracy, parameter count, computational complexity, and measured inference throughput. The results characterize robustness--efficiency trade-offs and support model selection against different computational constraints. We release the corruption definitions, severity configurations, evaluation framework, and associated results to support reproducible comparisons.
\end{enumerate}

The remainder of the paper describes the dataset, benchmark construction, and evaluation protocol; presents the recognition, robustness, and efficiency results; discusses their implications and limitations; and concludes with directions for future research.

\section{Related Work}
\label{sec:related}

\subsection{Tamil Handwritten Character Recognition}
Tamil handwritten-character recognition has been studied using handcrafted features, conventional classifiers, and deep learning methods~\cite{shanthi2010,Banumathi2011,bhattacharya2007,vijay2014,Prakash2018,kavitha2022}. The HPL isolated Tamil character dataset supported early recognition benchmarks, while the unconstrained Tamil Handwritten Character Database (uTHCD) provided a larger collection of handwritten samples representing diverse writing styles~\cite{Shaffi2021uthcd,iwfhr2006}. \citet{Shaffi2021uthcd} introduced uTHCD with 156 character classes and approximately 91,000 samples. Subsequent studies investigated deep-learning-based recognition and evaluated model performance on Tamil character datasets~\cite{kavitha2022new,kavitha2022}.

These studies provide important datasets and recognition baselines. However, their primary objective is character recognition under conventional evaluation settings, rather than systematic comparison under a standardized set of image corruptions. Although uTHCD contains natural variations in handwriting and image quality, these variations do not replace controlled evaluation across explicitly defined corruption types and severity levels.

\subsection{Corruption-Robustness Benchmarks and Degraded Handwriting}
ImageNet-C established a systematic framework for evaluating image classifiers under common corruptions at multiple severity levels, using mean Corruption Error (mCE) to summarize robustness~\cite{imagenet-c}. MNIST-C extended corruption-based evaluation to handwritten-digit classification, demonstrating the relevance of controlled degradation testing even for relatively simple visual recognition tasks~\cite{mnist-c}. Related work has also investigated robustness in scene-text detection, where image degradation can affect text localization and recognition~\cite{wu2021robustscenetext}. However, these benchmarks address different recognition tasks and do not directly establish robustness baselines for isolated Tamil handwritten characters.

Research on degraded handwriting in Indic scripts has addressed related but distinct problems. \citet{karki2018noisybangla} investigated pixel-level reconstruction and classification of noisy Bangla handwritten characters and numerals. \citet{chaudhury2022degraded} developed an OCR system for degraded Bangla documents, combining text-foreground segmentation with recognition. These studies address noise handling and degraded-document recognition, respectively, rather than providing a standardized, multi-corruption benchmark for comparing recognition architectures on Tamil handwritten characters.

\textbf{Research gap.} The literature reviewed for this study did not identify a benchmark specifically designed to evaluate isolated Tamil handwritten-character recognition across multiple corruption types and severity levels using a consistent evaluation protocol. To address this gap, we introduce uTHCD-C, which defines ten corruption types at five severity levels, including a stroke-thinning corruption tailored to handwritten glyphs. The benchmark provides a common evaluation framework for comparing nine recognition architectures and reporting their corruption robustness using mCE. Table~\ref{tab:table1} summarizes how the most relevant prior work relates to uTHCD-C.

\begin{table*}[h!]
\caption{Comparison of relevant prior work and the proposed uTHCD-C benchmark.}
\centering
\small
\setlength{\tabcolsep}{5pt}
\renewcommand{\arraystretch}{1.2}

\begin{tabular}{|p{3.0cm}|p{2.6cm}|p{3.0cm}|p{3.5cm}|}
\hline
\textbf{Study} & \textbf{Script and task} & \textbf{Degradation evaluation} & \textbf{Role relative to uTHCD-C} \\ \hline

\citet{Shaffi2021uthcd} &
Tamil, isolated HCR & Clean images only & Provides the uTHCD dataset that uTHCD-C extends \\ \hline

\citet{kavitha2022new} &
Tamil, isolated HCR & Clean images only & Recognition baseline on uTHCD without corruption evaluation \\ \hline

\citet{kavitha2022} &
Tamil, isolated HCR & Clean images only & CNN recognition baseline without corruption evaluation \\ \hline

\citet{karki2018noisybangla} &
Bangla, characters and numerals & Pixel-level noise (reconstruction and classification) & Addresses noise handling for a different script, not a benchmark \\ \hline

\citet{chaudhury2022degraded} &
Bangla, degraded-document OCR & Degraded scanned documents & Document-level OCR, not an architecture-level robustness benchmark \\ \hline

\textbf{This work (uTHCD-C)} &
Tamil, isolated HCR & Ten corruption types $\times$ five severity levels & Standardized multi-corruption benchmark comparing nine architectures via mCE \\ \hline

\end{tabular}

\label{tab:table1}
\end{table*}

\section{Methodology}
\label{sec:methodology}
\subsection{Dataset and Preprocessing}
We use the unconstrained Tamil Handwritten Character Database (uTHCD)~\cite{Shaffi2021uthcd}, which contains 90,950 handwritten character images from 156 Tamil character classes and 753 writers. The native image resolution is $64 \times 64$ pixels with three color channels. All images are resized to $224 \times 224$ pixels to accommodate the input requirements of the evaluated ImageNet-pretrained architectures. Table~\ref{tab:dataset} summarizes the dataset.

\begin{table}[h]
\caption{Summary of the uTHCD dataset.}
\centering
\footnotesize
\setlength{\tabcolsep}{5pt}
\begin{tabular}{@{}ll@{}}
\toprule
\textbf{Characteristic} & \textbf{Value} \\
\midrule
Dataset & uTHCD \\
Script & Tamil \\
Number of samples & 90,950 \\
Number of classes & 156 \\
Number of writers & 753 \\
Native image resolution & $64 \times 64 \times 3$ \\
\bottomrule
\end{tabular}
\label{tab:dataset}
\end{table}

The training pipeline uses random rotation within $\pm 7^\circ$, horizontal flipping, and color jitter, followed by normalization with the ImageNet channel means and standard deviations. The same preprocessing convention is maintained across architectures to facilitate comparison. Random augmentations are applied during training, whereas evaluation uses the unaugmented images or their deterministically generated corrupted counterparts.

\subsection{Recognition Architectures and Training}
We evaluate nine architectures spanning conventional convolutional networks and transformer-based or modern convolutional designs. The evaluated models are VGG16-BN~\cite{simonyan2015vgg}, GoogLeNet~\cite{Szegedy_2015_CVPR}, EfficientNet-B0~\cite{pmlr-v97-tan19a}, RegNetX-400MF~\cite{Radosavovic_2020_CVPR}, MnasNet-0.5~\cite{Tan_2019_CVPR}, ShuffleNet V2-0.5~\cite{ma2018shufflenetv2}, SqueezeNet 1.0~\cite{iandola2016squeezenet}, ConvNeXt-Tiny~\cite{liu2022convnet}, and Swin-T~\cite{liu2021Swin}. The final classification layers are adapted to the 156 Tamil character classes. VGG16-BN serves as the reference model for corruption-error normalization and provides a baseline for comparison with earlier Tamil HCR studies. Table~\ref{tab:models} lists each model with its exact \texttt{torchvision} variant.

Transfer learning has been applied to handwritten-character recognition across Indic scripts, including Tamil by \citet{gayathri2022tamil}, Devanagari by \citet{aneja2019devanagari}, and Telugu and Kannada by \citet{sahoo2020offline}. Surveys of offline handwritten Indic-script recognition also document the range of datasets and recognition approaches developed across these scripts~\citep{sharma2020survey, singh2023indic}. Following these precedents, we initialize all nine architectures with ImageNet-pretrained weights and fine-tune them for the 156-class Tamil HCR task using the same training settings. This protocol provides a consistent basis for comparing their corruption robustness under a common fine-tuning regime.

All architectures are fine-tuned using AdamW~\cite{ilya2019adamw} with a learning rate of $3\times10^{-4}$, weight decay of $1\times10^{-4}$, and a batch size of 32. Training proceeds for up to 30 epochs, with early stopping based on a held-out validation set and a patience of seven epochs. The same training settings are used across architectures. Experiments are implemented using PyTorch 2.0~\cite{PyTorch2019} on an NVIDIA A100 GPU with 40~GB of memory.

\begin{table}[h!]
\caption{Evaluated architectures and their exact \texttt{torchvision} variants.}
\centering
\footnotesize
\setlength{\tabcolsep}{6pt}
\begin{tabular}{ll}
\toprule
\textbf{Model} & \textbf{TorchVision variant} \\
\midrule
VGG16-BN          & \texttt{vgg16\_bn} \\
GoogLeNet         & \texttt{googlenet} \\
EfficientNet-B0   & \texttt{efficientnet\_b0} \\
RegNetX-400MF     & \texttt{regnet\_x\_400mf} \\
MnasNet-0.5       & \texttt{mnasnet0\_5} \\
ShuffleNet V2-0.5 & \texttt{shufflenet\_v2\_x0\_5} \\
SqueezeNet 1.0    & \texttt{squeezenet1\_0} \\
ConvNeXt-Tiny     & \texttt{convnext\_tiny} \\
Swin-T            & \texttt{swin\_t} \\
\bottomrule
\end{tabular}
\label{tab:models}
\end{table}

\subsection{Parameter Count, Computational Complexity, and Inference Throughput}
\label{sec:efficiency_protocol}
We characterize the evaluated architectures using parameter count, computational complexity, and inference throughput. These metrics capture different aspects of computational requirements and are reported separately.

\paragraph{Parameter count.} We report the number of learnable parameters in each evaluated architecture, including the classification head adapted to the 156 Tamil character classes. Parameter count is calculated by summing the number of elements across the model's parameter tensors. It characterizes the number of learned parameters but is not equivalent to the serialized checkpoint size, which also depends on numerical precision and storage format.

\paragraph{Computational complexity.} We report multiply--accumulate operations (MACs) for a single forward pass with a $224\times224$ input, expressed in GMACs. The values are obtained using an operation-counting profiler on each evaluated model configuration. MACs characterize the arithmetic operations counted by the profiler and should not be interpreted as a direct measurement of execution time.

\paragraph{Inference throughput.} We measure inference throughput on an NVIDIA A100 GPU with 40~GB of memory, using FP32 precision and $224\times224$ inputs. Each model is placed in evaluation mode and all measurements are taken under disabled gradient tracking (\texttt{torch.no\_grad()}). Images are processed in batches of $128$. For each model we run $20$ warm-up batches to reach steady-state execution and then time $60$ consecutive batches; the device is synchronized with \texttt{torch.cuda.synchronize()} immediately before and after the timed region, which is measured with a monotonic wall-clock timer (\texttt{time.perf\_counter}). Throughput is computed as the total number of images processed in the timed region divided by its elapsed time, and is reported in images per second; it therefore reflects batched execution rather than single-image latency. Disk access and host-side preprocessing are excluded so that the measurements characterize model execution under the specified setup. Because throughput depends on the hardware, numerical precision, batch size, software implementation, and timing procedure, the reported values are specific to this configuration, and the measurement script is released with the benchmark.

Parameter count, GMACs, and throughput are interpreted as complementary measurements rather than interchangeable indicators of efficiency. We use these measurements to compare the evaluated architectures under a common experimental setup; the resulting throughput values should not be treated as hardware-independent deployment estimates. The measurements are reported in Section~\ref{sec:efficiency_results}.

\subsection{Corruption Suite}
\label{sec:corruption_suite}
uTHCD-C comprises ten corruption types, each evaluated at five severity levels. The suite includes nine corruption types adapted from established corruption-benchmarking conventions~\cite{imagenet-c,mnist-c} and a handwriting-specific stroke-thinning operator. The corruptions represent noise, blur, geometric or digital degradation, and changes in stroke width.

The noise group comprises Gaussian, shot, and impulse noise. Gaussian and defocus blur represent focus-related degradation. Elastic deformation, pixelation, contrast reduction, and scaling simulate geometric distortions and digital image degradation. Stroke thinning addresses the loss of stroke width that can alter the distinguishing structure of handwritten glyphs.

\begin{figure}[t]
    \centering
    \includegraphics[width=1\linewidth]{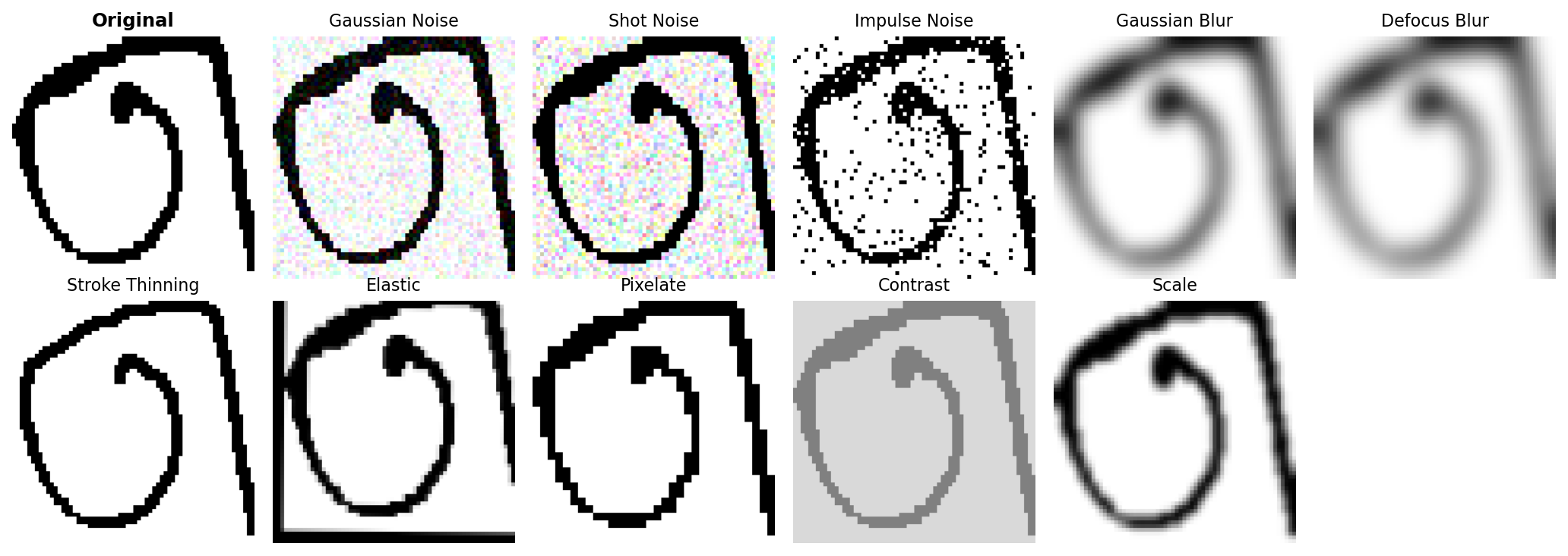}
    \caption{The ten corruption types in uTHCD-C applied to a single representative Tamil character at a fixed severity level (severity~3). Stroke thinning is the handwriting-specific operator; the remaining nine corruptions are adapted from established corruption benchmarks.}
    \label{fig:corruption_examples}
\end{figure}

For stroke thinning, each image is binarized and reduced to a single-pixel skeleton using Zhang--Suen thinning~\cite{zhang1984thinning}. The skeleton is then dilated with a square structuring element whose size controls the resulting stroke width. Smaller kernels produce thinner strokes. Table~\ref{tab:corruption_params} specifies the parameter values used at each severity level for all ten corruptions, and Figure~\ref{fig:corruption_examples} shows a representative Tamil character under each corruption at a single fixed severity level.

All corruptions are applied to the native $64 \times 64$ images before resizing them to the network input resolution. Noise magnitudes are specified on the $0$--$255$ pixel-intensity scale, and clean and corrupted images undergo the same subsequent preprocessing. Corrupted images are generated deterministically using random seeds associated with sample indices, ensuring that all architectures are evaluated on identical inputs. The corruption implementations and severity configurations are provided with the benchmark to support reproducible evaluation. The two mildest \emph{scale} levels (target sizes of $64$ pixels or greater) introduce relatively little degradation. For \emph{stroke thinning}, severity levels 4 and 5 use the same re-dilation kernel size of $1$ and therefore produce identical transformations, resulting in four distinct parameter settings across the five nominal severity levels.

\begin{table*}[t]
\centering
\caption{Corruption types and parameter settings for the five nominal severity levels. Parameter effects depend on the corruption operator: larger Gaussian-noise, impulse-noise, blur, and elastic-deformation parameters produce stronger corruption, whereas smaller shot-noise scales, contrast factors, and stroke-thinning re-dilation kernels produce stronger degradation. Pixelation uses the listed downsampling factor before restoring the image to its original resolution. For scale corruption, the listed intermediate target sizes are followed by restoration to the native $64 \times 64$ resolution; target sizes of 128 and 64 pixels do not constitute downsampling from the native resolution. Stroke-thinning levels 4 and 5 use the same kernel size of 1 and therefore produce identical transformations.}
\label{tab:corruption_params}
\begin{tabular}{llccccc}
\toprule
\textbf{Corruption} & \textbf{Parameter} & \textbf{s1} & \textbf{s2} & \textbf{s3} & \textbf{s4} & \textbf{s5} \\
\midrule
Gaussian noise  & std.\ dev.\ $\sigma$       & 10   & 20   & 30   & 40   & 60   \\
Shot noise      & Poisson scale              & 60   & 40   & 25   & 15   & 8    \\
Impulse noise   & corruption prob.           & 0.03 & 0.06 & 0.12 & 0.18 & 0.25 \\
Gaussian blur   & radius                     & 0.7  & 1.5  & 2.5  & 3.5  & 5.0  \\
Defocus blur    & radius                     & 1    & 2    & 3    & 4    & 6    \\
Stroke thinning & re-dilation kernel         & 9    & 7    & 3    & 1    & 1    \\
Elastic         & max.\ displacement (px)    & 1    & 2    & 3    & 4    & 6    \\
Pixelate        & downsampling factor        & 0.9  & 0.7  & 0.5  & 0.33 & 0.2  \\
Contrast        & contrast factor            & 0.7  & 0.5  & 0.35 & 0.2  & 0.1  \\
Scale           & intermediate target size (px) & 128  & 64   & 32   & 16   & 8    \\
\bottomrule
\end{tabular}
\end{table*}

\subsection{Evaluation Protocol and Robustness Metric}
\label{sec:eval_protocol}
The primary evaluation uses five-fold stratified cross-validation over the full uTHCD dataset. In each fold, a fresh model is fine-tuned from ImageNet-pretrained weights, and the held-out fold is evaluated under the clean condition and all ten corruption types at five severity levels. Each sample is held out for evaluation once across the five folds. Corruption robustness is summarized across folds using the mean and standard deviation.

We additionally evaluate generalization to unseen writers using a writer-independent partition. Writer identifiers are used to ensure that no writer contributes samples to both the training and test sets. The experiment uses a 70:30 partition with 226 of the 753 writers assigned to the held-out set; all 156 character classes are represented in the test set. Three independent partitions, using seeds 1, 42, and 123, are evaluated to assess the consistency of the findings.

Robustness is measured using the mean Corruption Error (mCE), following the ImageNet-C evaluation framework~\cite{imagenet-c}. Let $E_{m,c,s}$ denote the classification error of model $m$ under corruption $c$ at severity $s$, and let $E_{\mathrm{VGG16\text{-}BN},c,s}$ denote the corresponding error of the VGG16-BN reference model. For each corruption, the normalized corruption error is
\[
\mathrm{CE}_{m,c} =
\frac{\sum_{s=1}^{5} E_{m,c,s}}
{\sum_{s=1}^{5} E_{\mathrm{VGG16\text{-}BN},c,s}}.
\]
The aggregate corruption error is calculated by averaging over the ten corruption types:
\[
\mathrm{mCE}_{m} =
\frac{1}{10}\sum_{c=1}^{10}\mathrm{CE}_{m,c}.
\]
Lower mCE indicates lower corruption error relative to the reference model. The VGG16-BN reference has an mCE of 1 by construction. Clean-data accuracy is reported separately to contextualize the robustness results. Unless otherwise stated, the main corruption-robustness results are summarized over the five cross-validation folds.

\section{Results and Discussion}
\label{sec:results}

This section evaluates nine recognition architectures on uTHCD-C in terms of clean-data accuracy, computational requirements, corruption robustness, and writer-independent generalization. The results also examine the trade-off between corruption robustness and inference throughput. Unless otherwise specified, recognition and corruption-robustness results are reported as means over five-fold stratified cross-validation.

\subsection{Clean-Data Performance}

Table~\ref{tab:clean} reports clean-data accuracy under five-fold stratified cross-validation. All nine architectures achieve accuracies above $95\%$, with ConvNeXt obtaining the highest accuracy ($97.82\%$), followed closely by EfficientNet ($97.77\%$) and RegNet ($97.73\%$). SqueezeNet achieves the lowest accuracy ($95.74\%$).

\begin{table}[h!]
\centering
\caption{Clean-data recognition accuracy under 5-fold stratified cross-validation (mean\,$\pm$\,standard deviation over five folds, sorted by accuracy). The reference model is shown in bold.}
\label{tab:clean}
\begin{tabular}{lc}
\toprule
\textbf{Model} & \textbf{Clean accuracy (\%)} \\
\midrule
ConvNeXt      & 97.82\,$\pm$\,0.13 \\
EfficientNet  & 97.77\,$\pm$\,0.11 \\
RegNet        & 97.73\,$\pm$\,0.19 \\
Swin          & 97.57\,$\pm$\,0.14 \\
ShuffleNet    & 97.27\,$\pm$\,0.14 \\
MnasNet       & 97.11\,$\pm$\,0.24 \\
\textbf{VGG16-BN}  & \textbf{97.00\,$\pm$\,0.13} \\
GoogLeNet     & 96.33\,$\pm$\,0.68 \\
SqueezeNet    & 95.74\,$\pm$\,0.24 \\
\bottomrule
\end{tabular}
\end{table}

\subsection{Parameter Count, Computational Complexity, and Inference Throughput}
\label{sec:efficiency_results}

Table~\ref{tab:efficiency} summarizes the parameter count, computational complexity, and measured inference throughput of the evaluated architectures. Parameter counts range from $0.50$M for ShuffleNet to $134.9$M for VGG16-BN, while computational complexity ranges from $0.04$ to $15.48$ GMACs per forward pass. Measured throughput ranges from $1{,}072$ images/s for Swin to $30{,}591$ images/s for ShuffleNet under the evaluated A100 FP32 configuration.

\begin{table}[h!]
\centering
\caption{Resource characteristics of the evaluated architectures: parameter count (classifier resized to 156 classes), computational complexity (GMACs for one $224\times224$ forward pass), and inference throughput (images/s, measured on an NVIDIA A100 in FP32 at $224\times224$; higher is faster). Clean accuracy (5-fold mean) is repeated for reference. Sorted by parameter count. The reference model is shown in bold.}
\label{tab:efficiency}
\begin{tabular}{lcccc}
\toprule
\textbf{Model} & \textbf{Clean acc.\ (\%)} & \textbf{Params (M)} & \textbf{GMACs} & \textbf{Throughput (img/s)} \\
\midrule
ShuffleNet    & 97.27 & 0.50   & 0.04  & 30{,}591 \\
SqueezeNet    & 95.74 & 0.81   & 0.75  & 7{,}240  \\
MnasNet       & 97.11 & 1.14   & 0.10  & 12{,}750 \\
EfficientNet  & 97.77 & 4.21   & 0.38  & 5{,}110  \\
RegNet        & 97.73 & 5.16   & 0.41  & 10{,}242 \\
GoogLeNet     & 96.33 & 5.70   & 1.50  & 6{,}479  \\
Swin          & 97.57 & 27.64  & 4.50  & 1{,}072  \\
ConvNeXt      & 97.82 & 27.94  & 4.47  & 1{,}266  \\
\textbf{VGG16-BN}  & \textbf{97.00} & \textbf{134.9} & \textbf{15.48} & \textbf{1{,}498} \\
\bottomrule
\end{tabular}
\end{table}

The rankings differ across these measures. ShuffleNet has both the lowest computational complexity ($0.04$ GMACs) and the highest throughput, but computational complexity does not predict throughput in general: GoogLeNet ($1.50$ GMACs) attains higher throughput than EfficientNet ($0.38$ GMACs), and VGG16-BN, despite having the largest parameter count and computational complexity, achieves higher throughput than Swin and ConvNeXt. These results demonstrate substantial differences in computational requirements among architectures with similarly high clean-data accuracy.

\subsection{Aggregate Corruption Robustness}
\label{sec:robustness_mce}

Table~\ref{tab:mce_comparison} reports mean Corruption Error (mCE), normalized against the VGG16-BN reference using the ratio-of-sums definition described in Section~\ref{sec:eval_protocol}. Values are reported as means and standard deviations across the five folds; lower values indicate lower corruption error relative to the reference.

ConvNeXt achieves the lowest mCE ($0.75\pm0.03$), followed by Swin ($0.83\pm0.07$). EfficientNet ($0.91\pm0.16$) and GoogLeNet ($0.98\pm0.16$) also achieve values below the reference value of $1.00$. In contrast, SqueezeNet, RegNet, ShuffleNet, and MnasNet obtain mCE values of $1.10$, $1.25$, $1.55$, and $1.85$, respectively.

The results establish distinct corruption-robustness baselines for the nine architectures. In particular, MnasNet and ShuffleNet exhibit substantially higher aggregate corruption error than ConvNeXt and Swin, despite achieving clean-data accuracies above $97\%$.

\begin{table}[h!]
\centering
\caption{Aggregate corruption robustness under 5-fold stratified cross-validation, measured by mCE (VGG16-BN reference, lower is better; mean\,$\pm$\,standard deviation over five folds, sorted by mCE). The reference model is shown in bold.}
\label{tab:mce_comparison}
\begin{tabular}{lc}
\toprule
\textbf{Model} & \textbf{mCE} \\
\midrule
ConvNeXt      & 0.75\,$\pm$\,0.03 \\
Swin          & 0.83\,$\pm$\,0.07 \\
EfficientNet  & 0.91\,$\pm$\,0.16 \\
GoogLeNet     & 0.98\,$\pm$\,0.16 \\
\textbf{VGG16-BN}  & \textbf{1.00\,$\pm$\,0.00} \\
SqueezeNet    & 1.10\,$\pm$\,0.15 \\
RegNet        & 1.25\,$\pm$\,0.16 \\
ShuffleNet    & 1.55\,$\pm$\,0.22 \\
MnasNet       & 1.85\,$\pm$\,0.33 \\
\bottomrule
\end{tabular}
\end{table}

\subsection{Performance under Individual Corruptions}

Aggregate mCE summarizes performance across corruption types but can obscure differences under particular degradations. Table~\ref{tab:gauss} compares clean-data accuracy with accuracy under Gaussian noise at a standard deviation of $30$, illustrating the variation between architectures under a specific corruption condition.

ConvNeXt retains $97.81\%$ accuracy under this corruption, compared with $97.82\%$ on clean data. Swin retains $97.57\%$, while EfficientNet decreases from $97.77\%$ to $84.69\%$ and RegNet from $97.73\%$ to $68.35\%$. The corresponding accuracies for ShuffleNet and MnasNet fall to $11.16\%$ and $0.66\%$, respectively.

\begin{table}[h!]
\centering
\caption{Clean-data accuracy and accuracy under Gaussian noise ($\sigma=30$), 5-fold mean (\%). Models with similar clean accuracy differ sharply under additive noise.}
\label{tab:gauss}
\begin{tabular}{lcc}
\toprule
\textbf{Model} & \textbf{Clean (\%)} & \textbf{Gaussian $\sigma=30$ (\%)} \\
\midrule
ConvNeXt      & 97.82 & 97.81 \\
Swin          & 97.57 & 97.57 \\
VGG16-BN      & 97.00 & 96.44 \\
GoogLeNet     & 96.33 & 94.71 \\
SqueezeNet    & 95.74 & 94.91 \\
EfficientNet  & 97.77 & 84.69 \\
RegNet        & 97.73 & 68.35 \\
ShuffleNet    & 97.27 & 11.16 \\
MnasNet       & 97.11 & 0.66 \\
\bottomrule
\end{tabular}
\end{table}

\begin{figure}[h]
    \centering
    \includegraphics[width=1\linewidth]{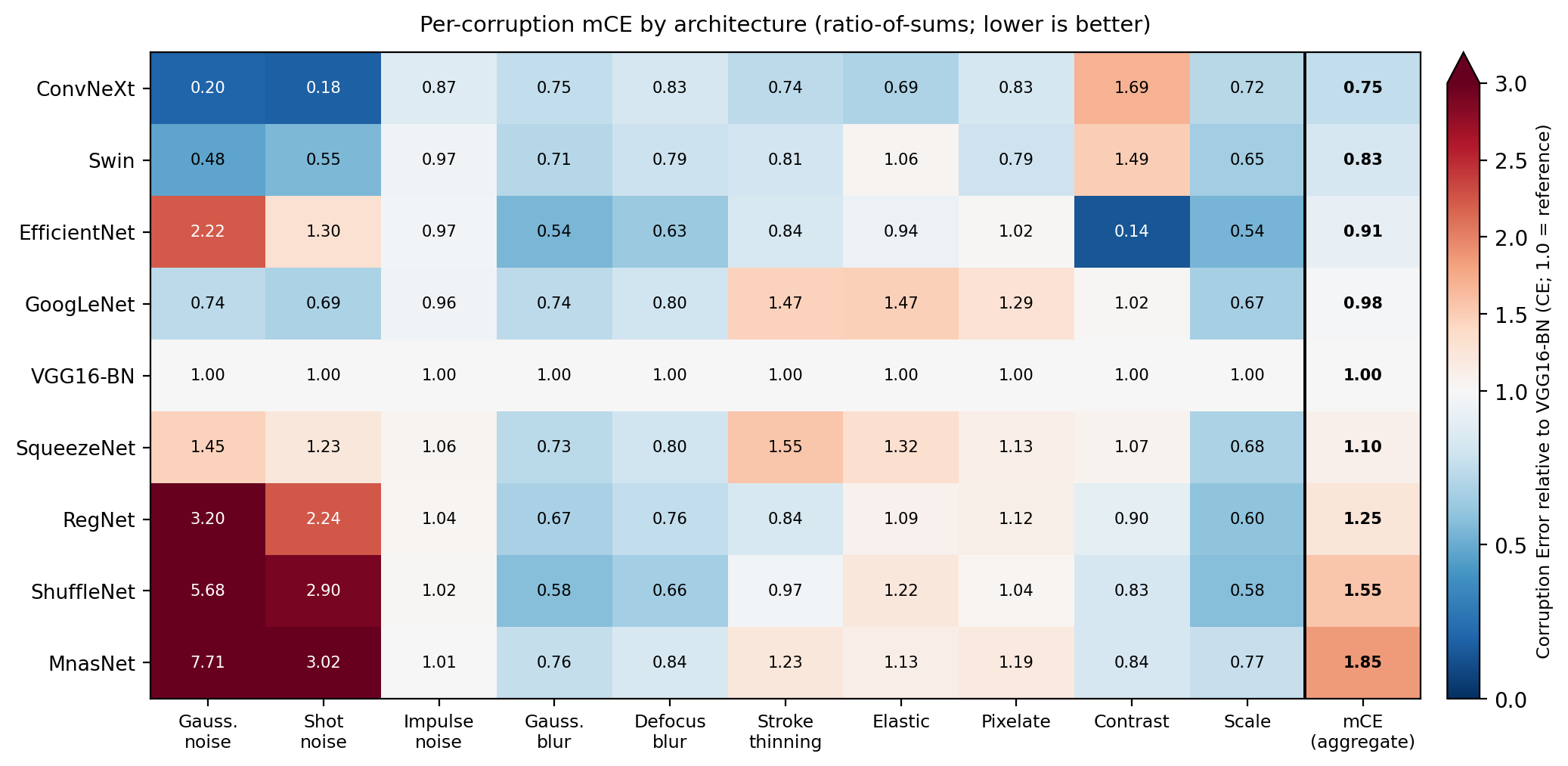}
    \caption{Per-corruption mCE by architecture (5-fold mean, ratio-of-sums definition; VGG16-BN $=1.0$, lower is better). Rows are ordered by aggregate mCE and the rightmost column repeats the aggregate value. The largest cross-architecture differences occur under Gaussian and shot noise, with additional variation under several other corruption types. The color scale is capped at $3.0$; cells exceeding this value are shown saturated, and the annotations report the exact mCE.}
    \label{fig:mce_heatmap}
\end{figure}

Figure~\ref{fig:mce_heatmap} presents the per-corruption mCE values across all architectures. The largest differences in relative corruption error occur under Gaussian and shot noise. The per-corruption accuracy results in Appendix~\ref{app:percorr} additionally show that impulse noise produces low absolute accuracy across architectures. Together, these results identify the degradation conditions associated with the largest performance differences and complement the aggregate robustness ranking.

\subsection{Writer-Independent Evaluation}

Table~\ref{tab:writer_indep} compares corruption robustness under a fixed writer-dependent $70{:}30$ split and three writer-independent partitions, using seeds $1$, $42$, and $123$. The writer-independent protocol ensures that writers represented in the test set are excluded from training.

ConvNeXt maintains an mCE of $0.81\pm0.03$, while Swin achieves $0.92\pm0.05$. MnasNet and ShuffleNet exhibit the highest writer-independent mCE values, at $2.65\pm0.59$ and $2.15\pm0.48$, respectively. Both models also show higher mCE under the writer-independent protocol than under the fixed writer-dependent split.

The results preserve the broad distinction between the more robust architectures and those with higher corruption error, while showing that the magnitude of the measured error depends on the evaluation protocol. The writer-independent results therefore complement the five-fold evaluation by characterizing robustness on writers excluded from training.

\begin{table}[h!]
\centering
\caption{Writer-independent corruption robustness (mCE, VGG16-BN reference, lower is better). ``Writer-dependent'' is the single fixed $70{:}30$ split; ``writer-independent'' is the mean\,$\pm$\,standard deviation over three writer-disjoint partitions (seeds $1$, $42$, $123$). Sorted by writer-independent mCE; the reference model is shown in bold.}
\label{tab:writer_indep}
\begin{tabular}{lcc}
\toprule
\textbf{Model} & \textbf{Writer-dependent mCE} & \textbf{Writer-independent mCE} \\
\midrule
ConvNeXt      & 0.81 & 0.81\,$\pm$\,0.03 \\
Swin          & 0.85 & 0.92\,$\pm$\,0.05 \\
\textbf{VGG16-BN}  & \textbf{1.00} & \textbf{1.00\,$\pm$\,0.00} \\
GoogLeNet     & 0.97 & 1.13\,$\pm$\,0.24 \\
EfficientNet  & 0.88 & 1.14\,$\pm$\,0.11 \\
SqueezeNet    & 1.13 & 1.32\,$\pm$\,0.09 \\
RegNet        & 1.38 & 1.43\,$\pm$\,0.11 \\
ShuffleNet    & 1.61 & 2.15\,$\pm$\,0.48 \\
MnasNet       & 2.08 & 2.65\,$\pm$\,0.59 \\
\bottomrule
\end{tabular}
\end{table}

\subsection{Robustness--Efficiency Trade-offs}
\label{sec:tradeoffs}

\begin{figure}[h]
    \centering
    \includegraphics[width=0.8\linewidth]{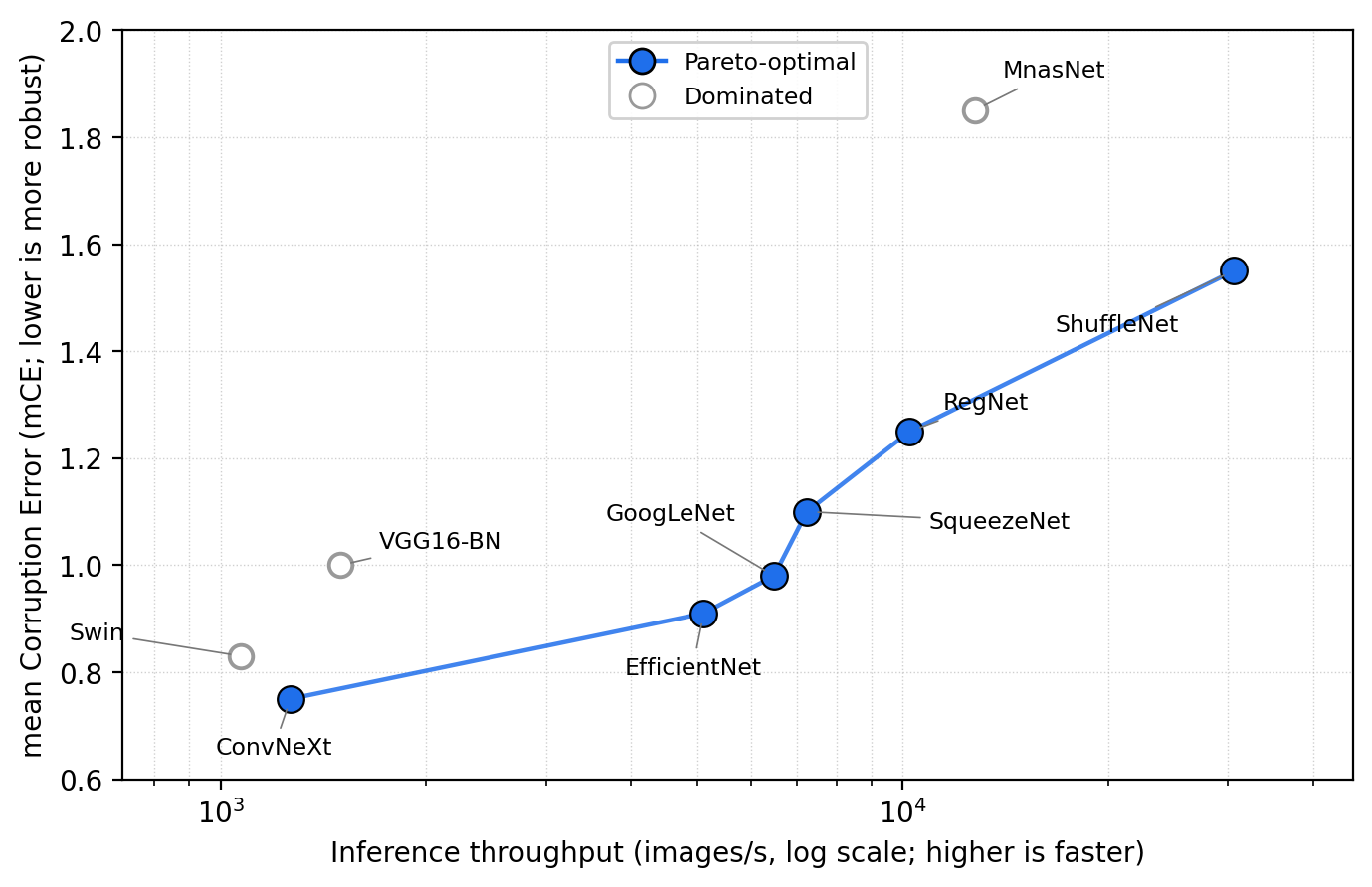}
    \caption{Pareto frontier of inference throughput and corruption robustness. Each point represents one evaluated architecture. Higher throughput and lower mCE are preferable. Pareto-optimal models are those for which no other evaluated architecture is simultaneously faster and more robust. The frontier is defined over these two criteria only.}
    \label{fig:pareto}
\end{figure}

To examine the relationship between corruption robustness and inference speed, we analyze the Pareto frontier defined by throughput and mCE. Throughput is maximized and mCE is minimized. An architecture is Pareto-optimal if no other evaluated architecture is both faster and more robust under these two criteria. Figure~\ref{fig:pareto} plots the evaluated architectures in throughput--mCE space, identifying the Pareto-optimal models and those dominated on both criteria.

Six architectures lie on the resulting frontier: ConvNeXt, EfficientNet, GoogLeNet, SqueezeNet, RegNet, and ShuffleNet. Ordered by increasing throughput, their mCE values increase from $0.75$ to $1.55$, describing a trade-off between measured inference speed and aggregate corruption robustness.

The remaining three architectures---Swin, VGG16-BN, and MnasNet---are dominated in this two-dimensional comparison. ConvNeXt is both faster and more robust than Swin; EfficientNet is both faster and more robust than VGG16-BN; and ShuffleNet is both faster and more robust than MnasNet.

The frontier identifies different operating points rather than a universally preferable architecture. ConvNeXt achieves the lowest mCE, whereas ShuffleNet achieves the highest throughput but has a substantially higher mCE. The appropriate choice therefore depends on the relative importance of corruption robustness and inference speed. Parameter count, computational complexity, and clean-data accuracy provide additional selection criteria, but they are not incorporated into this two-dimensional Pareto analysis.

\subsection{Summary}

The evaluation establishes corruption-robustness baselines for nine Tamil HCR architectures, characterizing substantial differences in aggregate robustness, corruption-specific failures, and writer-independent performance. The accompanying efficiency measurements and throughput--mCE Pareto analysis provide complementary evidence for comparing architectures under different computational and robustness requirements.

\section{Conclusion}
\label{sec:conclusion}
This study introduced uTHCD-C, a corruption-robustness benchmark for isolated Tamil handwritten character recognition based on the unconstrained Tamil Handwritten Character Database (uTHCD). The benchmark comprises ten corruption types evaluated at five severity levels, including a handwriting-specific stroke-thinning corruption. Using five-fold stratified cross-validation and writer-independent testing, we established comparative robustness baselines for nine convolutional and transformer-based architectures and quantified corruption robustness using mean Corruption Error (mCE).

The results characterize substantial differences in corruption robustness across architectures with similarly high clean-data accuracy. ConvNeXt and Swin achieved the lowest aggregate mCE values ($0.75$ and $0.83$, respectively), whereas ShuffleNet and MnasNet exhibited higher values ($1.55$ and $1.85$, respectively). Corruption-specific evaluation identified pronounced performance degradation under additive noise, while writer-independent testing showed that the principal differences in robustness persisted when evaluation was restricted to writers unseen during training.

The accompanying analysis of parameter count, computational complexity, and inference throughput extends the benchmark's utility to practical model comparison. The throughput--mCE Pareto analysis identified six non-dominated architectures, representing different trade-offs between inference speed and corruption robustness. These results provide a basis for selecting recognition models according to application-specific requirements rather than relying on a single performance measure. The efficiency measurements complement the benchmark's central contribution by characterizing the computational requirements associated with the observed robustness differences. By providing standardized corruption configurations, evaluation protocols, and comparative results, uTHCD-C establishes a reproducible foundation for further research on corruption robustness in Tamil handwritten-character recognition.

This study has limitations. The benchmark applies synthetic corruptions to a single dataset of isolated handwritten characters; therefore, the results do not establish how the evaluated architectures will perform on naturally degraded documents or other Tamil handwriting datasets. Inference throughput was measured under a specific hardware, precision, and implementation configuration, and the resulting comparisons may differ under other execution environments. Future work should investigate the relationship between synthetic and real-world degradation, evaluate additional handwriting datasets, and examine whether the observed robustness patterns extend to other scripts and recognition tasks.

\appendix
\section{Per-Corruption Accuracy}
\label{app:percorr}
Table~\ref{tab:percorr_summary} reports, for every architecture and every corruption, the accuracy (\%) averaged over the five severity levels (5-fold mean). These results complement the aggregate mCE of Table~\ref{tab:mce_comparison} and the per-corruption heatmap of Figure~\ref{fig:mce_heatmap}, which are computed from severity-specific errors normalized against the reference model. The largest accuracy reductions for MnasNet and ShuffleNet occur under Gaussian and shot noise, while impulse noise also yields low absolute accuracy across the evaluated architectures; under several other corruptions, including stroke thinning and elastic deformation, both models retain substantially higher accuracy. The complete per-severity tables for all ten corruptions are released with the benchmark code.

\begin{table*}[h!]
\centering
\caption{Accuracy (\%) under each corruption, averaged over the five severity levels (5-fold mean); higher is better. The noise columns (Gauss., Shot, Impulse) carry the largest differences across architectures. Full per-severity tables are available in the released benchmark.}
\label{tab:percorr_summary}
\footnotesize
\setlength{\tabcolsep}{4pt}
\begin{tabular}{lcccccccccc}
\toprule
\textbf{Model} & \textbf{Gauss.} & \textbf{Shot} & \textbf{Impulse} & \textbf{G.blur} & \textbf{Defoc.} & \textbf{Thin} & \textbf{Elastic} & \textbf{Pixel.} & \textbf{Contr.} & \textbf{Scale} \\
\midrule
VGG16-BN     & 86.1 & 66.0 & 11.1 & 22.9 & 19.1 & 95.8 & 94.0 & 70.5 & 77.2 & 57.2 \\
GoogLeNet    & 90.0 & 76.9 & 14.5 & 43.1 & 35.3 & 94.0 & 91.2 & 62.2 & 76.8 & 71.3 \\
EfficientNet & 71.5 & 58.2 & 13.4 & 58.5 & 49.1 & 96.5 & 94.4 & 70.1 & 96.8 & 77.0 \\
RegNet       & 59.3 & 26.8 & 7.9  & 48.1 & 38.7 & 96.5 & 93.5 & 67.1 & 79.4 & 74.3 \\
MnasNet      & 0.8  & 0.7  & 10.3 & 41.2 & 32.3 & 94.9 & 93.2 & 64.9 & 80.9 & 66.9 \\
ShuffleNet   & 25.7 & 4.6  & 9.4  & 55.6 & 46.6 & 96.0 & 92.7 & 69.3 & 81.1 & 75.3 \\
SqueezeNet   & 82.2 & 60.6 & 5.7  & 43.8 & 35.2 & 93.6 & 92.1 & 66.9 & 75.6 & 70.8 \\
ConvNeXt     & 97.3 & 93.7 & 22.9 & 42.0 & 32.6 & 97.0 & 95.9 & 75.7 & 61.5 & 69.0 \\
Swin         & 93.3 & 81.1 & 14.2 & 45.1 & 36.4 & 96.7 & 93.7 & 76.8 & 65.9 & 72.1 \\
\bottomrule
\end{tabular}
\end{table*}

\section*{Data and Code Availability}
uTHCD is available from its maintainers~\cite{Shaffi2021uthcd} and the HP Labs
isolated Tamil character set from its distributors~\cite{iwfhr2006}. The
corruption suite, the ratio-of-sums mCE implementation, the model-benchmarking
and aggregation scripts, and the raw per-model results behind every table and
figure are available at \url{https://github.com/pre-prakash/uTHCD-C-Bench}.

\bibliographystyle{plainnat}
\bibliography{references}

\end{document}